\documentclass{article}

\usepackage{spconf, graphicx,amssymb,lineno}
\usepackage{mathrsfs}
\usepackage{algpseudocode}
\usepackage{algorithmicx,algorithm}
\usepackage[usenames]{color}
\usepackage{graphicx,graphics,color,epsfig,subfigure,graphpap,rotate}
\usepackage{times, subfigure, graphicx, amsmath}
\usepackage{subfigure}
\usepackage{bm}
\usepackage{cite}
\usepackage{CJK}
\usepackage{multicol}
\usepackage{multirow}
\usepackage[colorlinks,
            linkcolor=blue,
            anchorcolor=blue,
            citecolor=blue]{hyperref}
\usepackage[font=small]{caption}
\usepackage{booktabs}
\usepackage{color}

\title{Information-bottleneck-based Behavior Representation Learning for Multi-agent Reinforcement learning}
\name{Yue~Jin$^{1}$ 
      \qquad Shuangqing~Wei$^{2}$ 
      \qquad Jian~Yuan$^{1}$
      \qquad Xudong~Zhang$^{1}$
      \thanks{This work was supported in part by the National Natural Science Foundation of China under Grant U20B2060.}
      }
\address{$^{1}$ Department of Electronic Engineering, Tsinghua University, Beijing, China \\
	     $^{2}$ School of Electrical Engineering and Computer Science, Louisiana State University, Baton Rouge, USA}

\begin{document}
\maketitle

\begin{abstract}
In multi-agent deep reinforcement learning, extracting sufficient and compact information of other agents is critical to attain efficient convergence and scalability of an algorithm.
In canonical frameworks, distilling of such information is often done in an implicit and uninterpretable manner, or explicitly with cost functions not able to reflect the relationship between information compression and utility in representation.
In this paper, we present Information-Bottleneck-based Other agents' behavior Representation learning for Multi-agent reinforcement learning (IBORM) to explicitly seek low-dimensional mapping encoder through which a compact and informative representation relevant to other agents' behaviors is established. 
%
IBORM leverages the information bottleneck principle to compress observation information, while retaining sufficient  information relevant to other agents' behaviors used for cooperation decision.
Empirical results have demonstrated that IBORM delivers the fastest convergence rate and the best performance of the learned policies, as compared with implicit behavior representation learning and explicit behavior representation learning without explicitly considering information compression and utility.
\end{abstract}


\section{Introduction}
\label{sec:intro}
Representation learning, which aims to learn informative and effective features of a task, is a key part of deep learning.
Naturally, deep reinforcement learning (DRL) is expected to benefit from the help of representation learning.
Many works \cite{lange2010deep, laskin2020curl, pacelli2020learning, franccois2019combined} have dedicated to boost DRL by learning a compact, discriminative or task-relevant representation from observations.
However, in multi-agent tasks, a good task-relevant representation also needs to be teammate-relevant or opponent-relevant.
It has been demonstrated in some works \cite{he2016opponent, jin2020stabilizing, foerster2017stabilising} that using independent DRL (ignore other agents' behaviors) may lead to unsatisfactory results.
Meanwhile, some works \cite{lowe2017multi, tesauro2004extending} indicate that inferring other agents' policies can improve cooperation between agents, but is prohibitively expensive for policies parameterized by deep neural networks.
These studies imply the demand for more efficient and effective representation learning of other agents in multi-agent DRL (MADRL).

Two core problems of representing other agents in MADRL are what to represent and how to combine the representation with MADRL.
Foerster et al. \cite{foerster2017stabilising} leverage low-dimensional fingerprints to represent other agents' policy changes, which forms succinct features, but is deficient in policy information completeness.
Jin et al. \cite{jin2020stabilizing} propose to represent other agents' behaviors implicitly using their positions at adjacent timesteps. 
However, the behavior representation is learned via MADRL in an implicit and uninterpretable manner.
He et al. \cite{he2016opponent} propose deep reinforcement opponent network (DRON) to learn representations of other agents' actions explicitly by leveraging other agents' actions as supervision signals.
However, the compactness and information utility of representation are not considered.

In this paper, we leverage information bottleneck principle \cite{tishby2015deep, tishby2000information} 
to learn an informative and compact representation relevant to other agents' behaviors to improve the performance of MADRL.
In particular,
we employ an encoder to extract features from each agent's positions at two adjacent timesteps, based on which a classifier is learned to estimate actions of each agent.
To filter out irrelevant information from observations and retain sufficient amount of information of other agents' actions used for cooperation decision, we follow the information bottleneck principle to minimize the mutual information between the representation and the observations, while maximizing the mutual information between the representation and other agents' actions.
To this end, we adopt a variational method \cite{belghazi2018mutual} to estimate the two mutual information and integrate this process into behavior representation learning.
We combine our proposed behavior representation learning method with our recent work, stabilized multi-agent deep Q learning (SMADQN) \cite{jin2020stabilizing} by multi-task learning and thereby the learned representation can also retain other information about the MADRL task in addition to other agents' actions. Experimental results demonstrate the superior performance of our proposed method compared to vanilla SMADQN and DRON-based SMADQN.

In summary, the main contributions of this paper are as follows:

1) We propose an information-bottleneck-based behavior representation learning method through which compact and informative features of other agents' behaviors are learned and exploited to facilitate MADRL.

2) We conduct extensive experiments in cooperative navigation tasks \cite{lowe2017multi, jin2020stabilizing, jin2019efficient}.
Experimental results demonstrate that compared to implicit behavior representation learning and the explicit behavior representation learning that does not consider information utility and compression, our method performs best in terms of both
learning speed and the success rates of the resulting policies.

\section{Method}
In this section, we first introduce the Markov game, SMADQN \cite{jin2020stabilizing} and DRON \cite{he2016opponent}.
Then, we present our method.

A Markov game with $N$ agents involves a set of states $s$, joint actions $(a_1,\cdots, a_N)$, transition probability function $p(s'|s,a_1,\cdots, a_N)$, and each agent's reward function $r_i(s,a_1,\cdots, a_N)$, $i\in [1,N]$.
At each timestep, each agent executes an action according to its policy $\pi_i$.
A problem of Markov game is to find the optimal policy $\pi_i^*$ for each agent so that
\footnotesize $\forall \pi_i, R_i(s^t, \pi_1^*, \cdots, \pi_i^*, \cdots, \pi_N^*)\geq R_i(s^t, \pi_1^*, \cdots, \pi_i, \cdots, \pi_N^*)$,
\normalsize where
\footnotesize $R_i(s^t, \pi_1, \cdots, \pi_N)=E[\sum\nolimits_{\tau=t}^T {{\gamma ^{\tau-t}}} {r_i(s^{\tau},a_1^{\tau}, \cdots, a_N^{\tau} )}]$
\normalsize denotes the expected total reward of agent $i$, $T$ is the time horizon, $\gamma\in[0,1]$ is a discount factor.
For convenience, we use $r_i^t$ to denote the reward of agent $i$ at timestep $t$.

SMADQN defines an extended action-value function $G$ for each agent to measure its expected total reward when it follows policy $\pi_{i}$. For agent $i$, $G$-function is defined as
\begin{equation}\label{action-value function}
\footnotesize
G_i^{\pi_{i}}(s^t,s_{-i}^t,s_{-i}^{t+1},a_i^t) =  Q_i^{\pi_{i}}(s^t,f(s_{-i}^t, s_{-i}^{t+1}),a_i^t),
\end{equation}
where $s^t$ represents global states, $s_{-i}^t$ and $s_{-i}^{t+1}$ represent states of agents except agent $i$ at two adjacent timesteps, $f$ is an action estimation function of other agents' actions. $G$-function is a composite function that incorporates the action estimation function into the original action-value function $Q$ \cite{sutton2018reinforcement}. 
An approximate extended Bellman equation for the optimal $G$-function is derived as:
\begin{equation}\label{Bellman_optimality_G}
\footnotesize
\begin{aligned}
&\mathbb{E}_{s_{-i}^{t+1}|s_{-i}^t,a_{-i}^t} G_i^*(s^t, s_{-i}^t, s_{-i}^{t+1},a^t_i) \approx\\  &\mathbb{E}_{s^{t+1}|s^t,a_i^t,a_{-i}^t}
\left[r_i^{t+1}+\gamma \mathop{\max}\limits_{a^{t+1}_i}G_i^*(s^{t+1},s_{-i}^t,s_{-i}^{t+1},a^{t+1}_i) \right].\\
\end{aligned}
\end{equation}
The optimal $G$-function is approximated by a neural network learned by minimizing the loss function given as:
\begin{equation}\label{loss_masddpg}
\footnotesize
\begin{aligned}
    &L=\mathbb{E}_{s^t, s^{t+1},a^t_i}\left[\left(r_i^{t+1}+\gamma \max\limits_{a_i^{t+1}} G_i(s^{t+1},s_{-i}^t,s_{-i}^{t+1},a^{t+1}_i)\right.\right.\\
    &\quad\quad\quad\quad \quad\quad \quad\quad \quad\quad  \left.\left.-G_i(s^t, s_{-i}^t. s_{-i}^{t+1},a^t_i)\right)^2\right],
\end{aligned}
\end{equation}

SMADQN learns action estimation function of other agents' actions implicitly, which may lead to trivial estimation performance and thereby cause limited performance of the resulting policies.

Instead of merging action estimation learning into  MADRL, DRON leverages other agents' actions as supervision signals and adopts supervised learning to learn action estimation.
It uses a classification network to estimate other agents' actions. The output of the last hidden layer of the network is used as other agents' action representation, and is fed into a decision network.
DRON can learn representations of other agents' behaviors explicitly.
However, it does not consider information compression and retention in the representation.

To facilitate and improve MADRL, extracting informative and compressed representation of other agents' behaviors is critical.
To this end, we propose Information-Bottleneck-based Other agents' behavior Representation learning for Multi-agent reinforcement learning (IBORM), which equips a behavior representation with the following  capabilities, a) to extract features of other agents' actions, b) to filter out irrelevant information while retaining sufficient information about the actions and other potentially helpful information about the task to facilitate MADRL.
Specifically, we implement our idea in SMADQN.
We replace the implicit action representation learning of SMADQN with explicit action representation learning.
An encoder is employed to learn the representation using other agents' states at adjacent timesteps as inputs.
The encoder's output is a low-dimensional feature vector of other agents' actions, from which a classifier can predict the actions.
The representation learning and SMADQN are combined by leveraging multi-task learning.

Additionally, to learn an informative and compressed representation, we leverage information bottleneck principle \cite{tishby2015deep, tishby2000information} to constrain the information contained in the representation.
To be specific, information bottleneck (IB) principle introduces an information theory principle for extracting an optimal representation $Z$ that captures the relevant information in a random variable $X$ about another correlated random variable $Y$ while minimizing the amount of irrelevant information, where $(Y,X,Z)$ forms a Markov chain, $Y\rightarrow X\rightarrow Z$.  Namely, finding the optimal representation function is formulated as minimizing the following Lagrangian
\begin{equation}\label{loss_IB}
\footnotesize
 \mathcal{L}(p(z|x)) = I(X; Z) - \kappa I(Z;Y),
\end{equation}
where $\kappa$ determines how much relevant information is contained in the representation.
Based on IB principle, we constrain the representation learning by the following terms
\begin{equation}\label{loss_IB_ours}
\footnotesize
\begin{aligned}
 &  \mathcal{L}(\alpha) \triangleq  I(\phi_{s_{-i,j}}; ENC_i^{\alpha}(\phi_{s_{-i,j}}) - \kappa I(ENC_i^{\alpha}(\phi_{s_{-i,j}});a_{-i,j}), \\
\end{aligned}
\end{equation}
where $\alpha$ denotes the parameters of the encoder, $a_{-i,j}$ and $\phi_{s_{-i,j}}$ denote the action and agent $i$'s observation of the $j$th agent other than agent $i$, respectively.

Overall, the loss function of IBORM is given as:
\begin{equation}\label{loss function}
    \footnotesize
    \begin{aligned}
    &L_i(\alpha, \beta, \theta)=J_i^{CE}(\alpha,\beta)+ \lambda_1 J_i^{DRL}(\alpha, \theta)+ \lambda_2 \mathcal{L}(\alpha),\\
    \end{aligned}
\end{equation}
where $J^{CE}_i(\alpha, \beta)$ denotes the cross-entropy between classifier’s output and each agent’s true action , $\alpha$ and $\beta$ are parameters of the encoder and the classifier.
$J_i^{DRL}(\alpha, \theta)$ denotes a modified loss function of SMADQN, defined as
\begin{equation}\label{drl loss}
    \footnotesize
    \begin{aligned}
    &J_i^{DRL}(\alpha, \theta) = \mathbb{E}_{\phi_{s_i}^t,a_i^t, \phi_{s_{-i,1}}^{t+1},\cdots,\phi_{s_{-i,N-1}}^{t+1}}\\ &\left[\left(y_i^t-G_i^{\theta}(\phi_{s_i}^t,ENC_i^{\alpha}(\phi_{s_{-i,1}}^{t+1}),
    \cdots,ENC_i^{\alpha}(\phi_{s_{-i,N-1}}^{t+1}),a_i^t)\right)^2\right],
    \end{aligned}
\end{equation}
where
\footnotesize $y_i^t=r_i^{t+1}+\gamma\mathop{\max}\nolimits_{a_i}G_i^{\theta_{tar}}(\phi_{s_i}^{t+1},ENC_i^{\alpha}(\phi_{s_{-i,1}}^{t+1}),
    \cdots,\\
    ENC_i^{\alpha}(\phi_{s_{-i,N-1}}^{t+1}),a_i)$.
\normalsize
$\theta$ and $\theta_{tar}$ are parameters of $G$ network and target network \cite{mnih2015human}, respectively.
Compared with (\ref{loss_masddpg}), we replace other agents' adjacent states with $N-1$ behavior representations.
For notation convenience, we rewrite (\ref{loss function}) as
\begin{equation}\label{loss function_new}
    \footnotesize
    \begin{aligned}
    &L_i(\alpha, \beta, \theta)=J_i^{CE}(\alpha,\beta)+ \lambda_1 J_i^{DRL}(\alpha, \theta)+  \\
    &\lambda_2 I(\phi_{s_{-i,j}}, ENC_i^{\alpha}(\phi_{s_{-i,j}})) - \lambda_3 I(ENC_i^{\alpha}(\phi_{s_{-i,j}}), a_{-i,j}),\\
    \end{aligned}
\end{equation}
where $\lambda_1, \lambda_2, \lambda_3$ are positive weights.
From the perspective of information utility, the first and the last terms of (\ref{loss function_new}) are for extracting sufficient information of other agents' behaviors.
The second term extracts relevant information of the task.
The third term filters out irrelevant information.
Compared to IBORM, SMADQN only uses the DRL-based term, where the behavior representation learning is implicitly contained. DRON uses the cross-entropy term but does not constrain the amount and utility of the information in the representation.

\begin{figure}[!t]
    \centering
    \includegraphics[width=0.6\columnwidth]{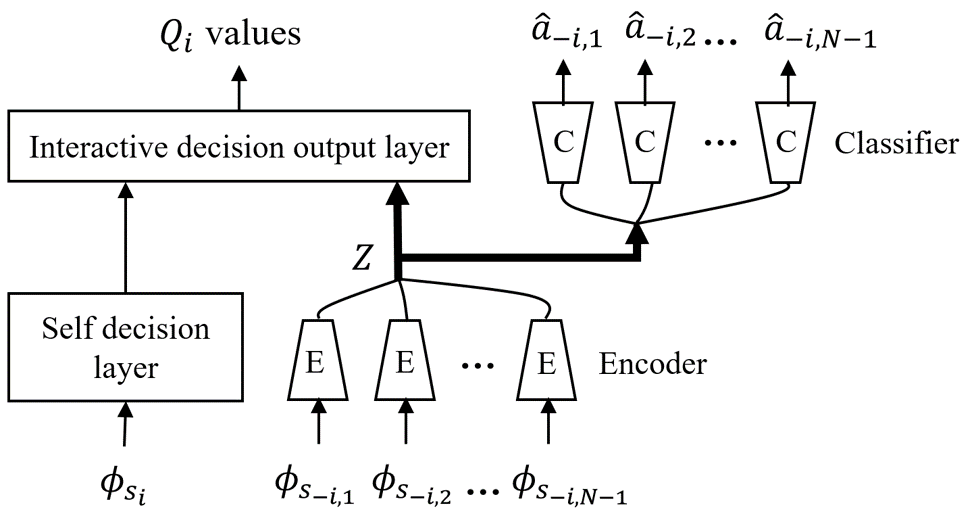}
    \captionsetup{font=footnotesize}
    \caption{Network architecture diagram of IBORM.}
    \vspace{-1mm}
    \label{fig.DiagramIBOR2}
\end{figure}

The network architecture of IBORM is shown in Fig.~\ref{fig.DiagramIBOR2}.
\begin{figure}[!]
    \centering
    \includegraphics[width=0.7\columnwidth]{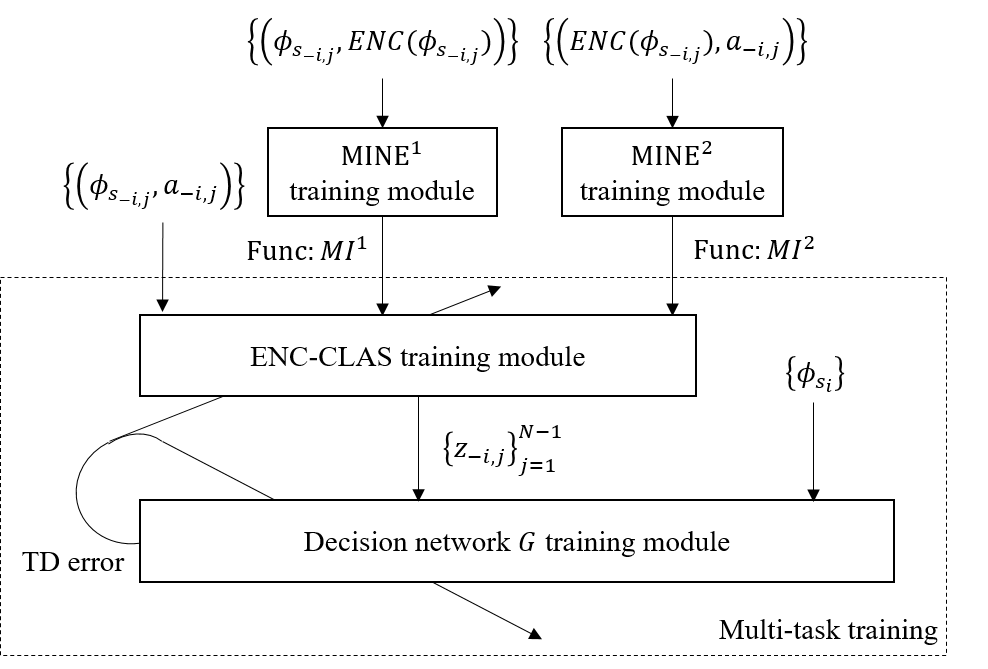}
    \captionsetup{font=footnotesize}
    \caption{Framework of IBORM algorithm.}
    \vspace{-1mm}
    \label{fig.AlgoFramework}
\end{figure}
An encoder-classifier is used to estimate each of other agents' actions.
The encoder is duplicated by $N-1$ times and generates bottleneck representations for $N-1$ other agents' actions, respectively.
Then, the representations are incorporated into a decision network to make interactive decisions.

To estimate the two mutual information terms in (\ref{loss function_new}), we employ the Mutual Information Neural Estimator (MINE) \cite{belghazi2018mutual} that estimates the mutual information between two variables $X$ and $Z$ as \footnotesize $\widehat{I(X,Z)}=\sup_{{\omega \in \Omega}} \mathbb{E}_{\mathbb{P}_{XZ}}[T_\omega(x,z)]-\log(\mathbb{E}_{\mathbb{P}_X \otimes \mathbb{P}_Z} [e^{T_\omega(x,z)}] )$ \normalsize by leveraging a trainable neural network $T_\omega$ with parameters $\omega$. Specifically, we use two networks corresponding to MINEs of the two mutual information terms in (\ref{loss function_new}). To integrate learning of MINEs and IBORM, we adopt an interlaced learning manner to update parameters of MINEs and IBORM alternately.
A framework of our algorithm is shown in Fig.~\ref{fig.AlgoFramework}.
The complete algorithm is shown in Algorithm \ref{algo}, where we denote
$\boldsymbol{\phi}_{s_{-i}}^t$ \footnotesize $ =[\phi_{s_{-i,1}}^{t},\cdots,\phi_{s_{-i,N-1}}^{t}]$
\normalsize
and
\footnotesize $\boldsymbol{a}_{-i}^t = [a_{-i,1}^t,\cdots,a_{-i,N-1}^t]$
\normalsize
for notation convenience.

\begin{algorithm}[!t]
\footnotesize
\captionsetup{font=small}
\caption{\footnotesize Stabilized multi-agent deep Q-learning with information-bottleneck-based other agents' behavior representation learning}
\begin{algorithmic}[1]
\raggedright
\For{agent $i=1$ to $N$}
    \State
    Initialize \\
    \qquad  networks \scriptsize $ENC_i^{\alpha}:\phi_{s_{-i,j}}\rightarrow z_{-i,j}$, $CLAS_i^{\beta}:z_{-i,j} \rightarrow \hat{a}_{-i,j}$, \\
    \qquad  $MINE_i^{\omega_1}(\phi_{s_{-i,j}}, z_{-i,j})$, $MINE_i^{\omega_2}(z_{-i,j},a_{-i,j})$, \\
    \qquad  $G_i^{\theta}(\phi_{s_i},ENC_i^{\alpha}(\phi_{s_{-i,1}}),\cdots,ENC_i^{\alpha}(\phi_{s_{-i,N-1}}),a_i)$,\\
    \qquad  \footnotesize target network \scriptsize $G_i^{\theta_{tar}}$ \footnotesize with \scriptsize $\theta_{tar}\leftarrow\theta$, \footnotesize replay buffer \scriptsize $\mathcal{D}_i$\\
\EndFor \footnotesize
\For{episode$=1$ to $Z$}
    \State Receive $\phi_{s_i}^1, \boldsymbol{\phi}_{s_{-i}}^1$ for each agent
    \For {t$=1$ to $T$}
        \State Execute action for each agent $i$:\\
        \scriptsize $a_i^t = \arg\max\limits_{a_i} G_i^{\theta}(\phi_{s_i}^t,ENC_i^{\alpha}(\phi_{s_{-i,1}}^t),\cdots,ENC_i^{\alpha}(\phi_{s_{-i,N-1}}^t),a_i)$
        \footnotesize
        \State Receive $r_i^{t+1}, \phi_{s_i}^{t+1}, \boldsymbol{\phi}_{s_{-i}}^{t+1}$ for each agent $i$
        \State Record data: \scriptsize $\mathcal{D}_i \leftarrow \mathcal{D}_i \bigcup \{(\phi_{s_i}^t,a_i^t,r_i^{t+1},\phi_{s_i}^{t+1},\boldsymbol{\phi}_{s_{-i}}^{t+1},\boldsymbol{a}_{-i}^t)\}$\\
        \qquad \quad ~ \footnotesize for each agent $i$
        \For {agent $i=1$ to $N$}
            \State Sample $M$ tuples \scriptsize \\
            \qquad \qquad \quad $\{(\phi_{s_i}^{\tau_k},a_i^{\tau_k},r_i^{\tau_k+1},\phi_{s_i}^{\tau_k+1},\boldsymbol{\phi}_{s_{-i}}^{\tau_k+1},\boldsymbol{a}_{-i}^{\tau_k})\}_{k=1}^M$ from $\mathcal{D}_i$ \footnotesize
                \State Compute \scriptsize~$\{z_{-i,j}^{\tau_k+1} = ENC_i^\alpha(\phi_{s_{-i,j}}^{\tau_k+1})\}_{
                j=1}^{N-1} ~ $\footnotesize 
                \State Compute $y_i^{\tau_k}$ by\\
                \scriptsize \quad \quad $y_i^{\tau_k}=r_i^{\tau_k+1}+\gamma\mathop{\max}\nolimits_{a_i}G_i^{\theta_{tar}}(\phi_{s_i}^{\tau_k+1},z_{-i,1}^{\tau_k+1},\cdots,z_{-i,N-1}^{\tau_k+1},a_i)$
            \State \footnotesize Update $\omega_1, \omega_2$ according to \cite{belghazi2018mutual} using SGD
            \State Update $\alpha, \beta, \theta$ to minimize (\ref{loss function_new}) using SGD
            \State Update target network with soft update rate $\eta$:\\
                \qquad \qquad ~~$\theta_{tar}\leftarrow\eta\theta+(1-\eta)\theta^{tar}$\\
        \EndFor
    \EndFor
\EndFor

\end{algorithmic}
\label{algo}
\end{algorithm}

\section{Experiments}
\begin{figure}[!t]
 \centering
\includegraphics[width=0.75\columnwidth]{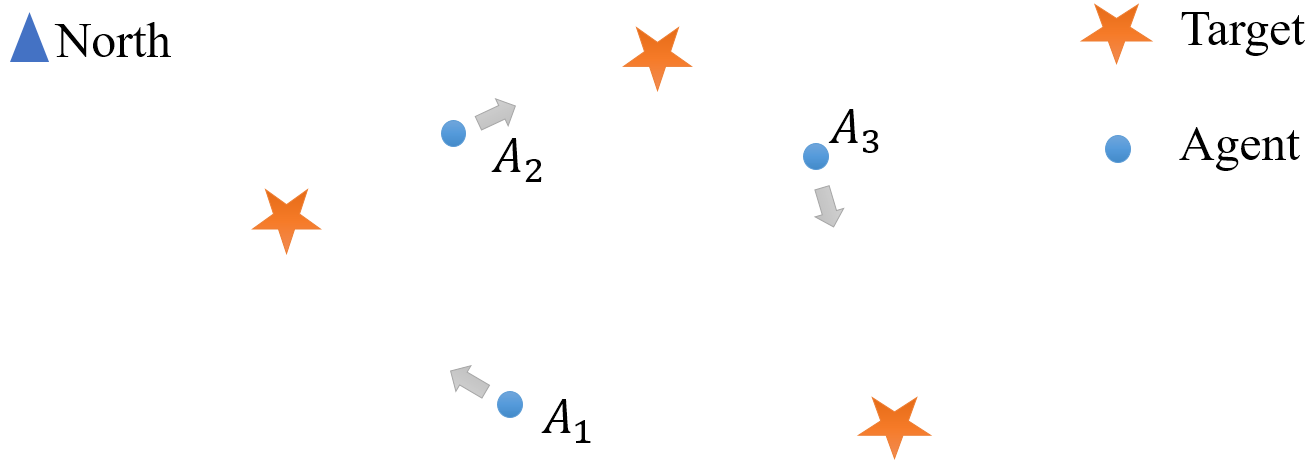}
\captionsetup{font=footnotesize}
\caption{Illustration of the cooperative navigation task involving three agents. }\vspace{-1mm}\label{fig:control scene}
\label{env}
\end{figure}

In this section, we evaluate IBORM in multi-agent cooperative navigation task with the same settings used in \cite{jin2020stabilizing}.
In this task, agents need to cooperate through motions to reach the same number of targets using the minimum time.
An example containing three agents and targets is illustrated in Fig.~\ref{env}.
At each timestep, each agent selects a target and move a fixed distance toward the target.
The action of an agent is defined as $a_i \in [1,N]$ that indicates the index of the target selected by it. Agents' speed is $1 \ \text{m}/\text{timestep}$.
The current observation of agent $i$, i.e. $\phi_{s_i}$, is composed of the current positions of other entities (targets and the other agents) and agent $i$'s last action.
An agent's observation about other agents' states, i.e.
$\phi_{s_{-i}}$, includes the positions of targets and the current and last positions of other agents, which are necessary for an agent to predict other agents' actions.
The size of the environment is $15\times15~m^2$.
The maximum episode length is $30$ timesteps.
Agents are homogeneous. They share a common policy and reward function.
The reward function is aligned with \cite{jin2020stabilizing}.

Network structure and hyperparameters of IBORM are as follows.
The decision network has two hidden layers containing $300$ and $200$ units, respectively.
The encoder and classifier both have two hidden layers. Encoder has $32$ and $16$ units in its two layers, respectively. Classifier has $16$ and $32$ units in its two layers, respectively.
Adam optimizer with a learning rate of 0.01 is applied to update parameters.
The size of the replay buffer is 1500, the batch size of SGD is 32, $\gamma=1$ and the target network is updated with a soft update rate \cite{lillicrap2015continuous} $\eta = 0.001$, which are all aligned with SMADQN.
$\lambda_1, \lambda_2$ and $\lambda_3$ are set as 10, 0.001 and 0.1 by grid search method.

We compare IBORM with SMADQN and a comparable DRON whose DRL loss are aligned with SMADQN.
We name the latter SMADRON.
For SMADQN,  
its network has two hidden layers containing the the same number of units as that in IBORM's decision network.
For SMADRON, its network structure is the same as IBORM. 
Hyperparameters used in SMADQN and SMADRON are the same as IBORM.

\begin{table}[t]
\renewcommand\arraystretch{1.1}
\scriptsize
\captionsetup{font=footnotesize}
\caption{Test results of different methods.}
\centering
\begin{tabular}{l l l l l l}
\toprule [1pt]
\multirow{2}*{Method} & \multicolumn{5}{c}{Success rate} \\
 & N=3 & N=4 & N=5 & N=6 &N=7 \\
\hline
SMADQN &98.2\%  &97.8\%  &96.1\%  &91.2\%  &0.0\% \\
SMADRON  &98.9\%  &96.9\%  &92.9\%  &93.5\%  &82.5\% \\
IBORM   &\textbf{99.3}\%  &\textbf{98.1}\%  &\textbf{97.1}\%  &\textbf{93.5}\%  &\textbf{87.8}\% \\
\bottomrule [1pt]
\end{tabular}
\label{success rate}
\end{table}

\begin{figure}[t]
\centering
\subfigure[]
{
    \begin{minipage}[t]{4cm}
	\centering
    \includegraphics[width=1.05\columnwidth]{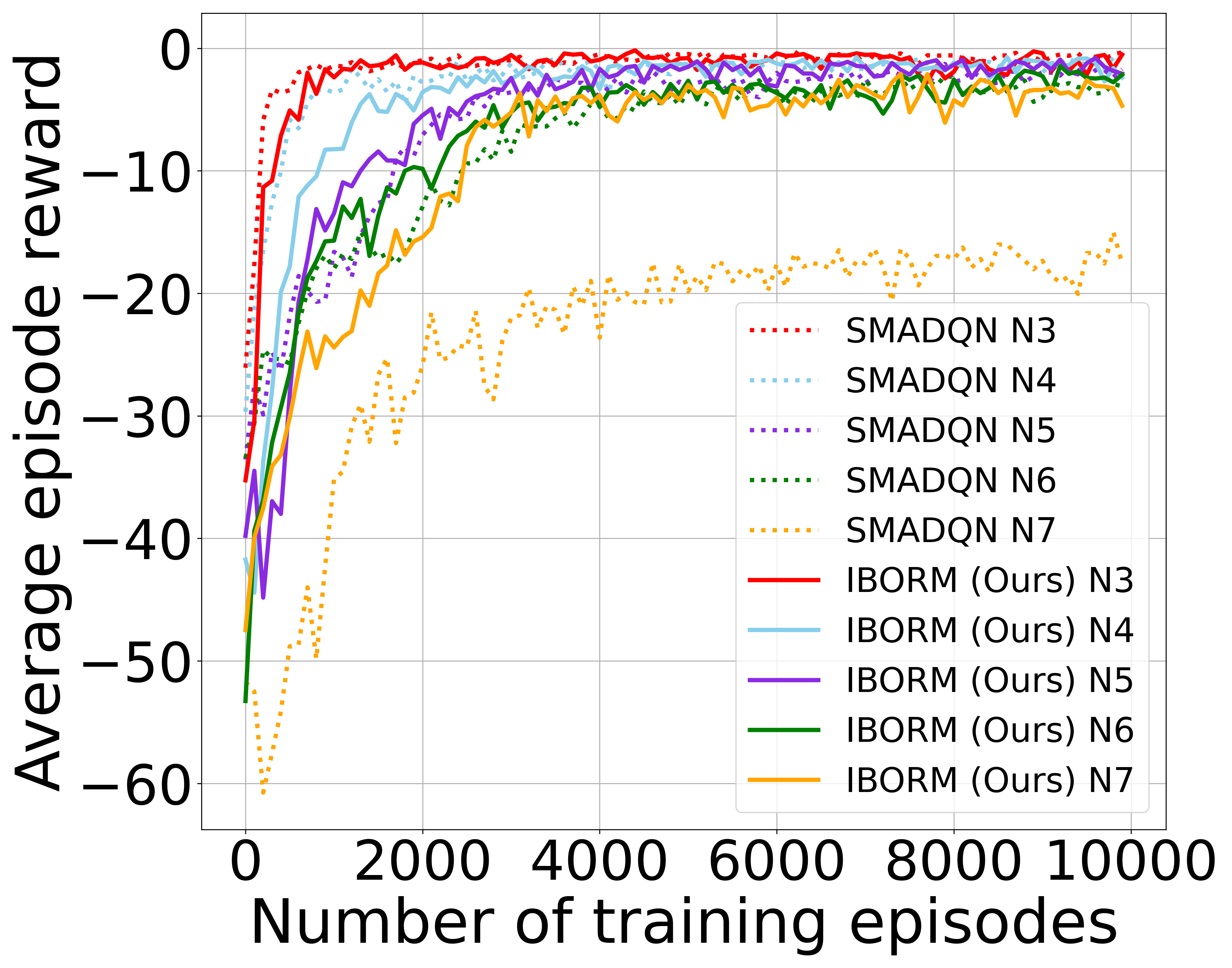}
    \end{minipage}
}
\subfigure[]
{
    \begin{minipage}[t]{4cm}
	\centering
    \includegraphics[width=1.05\columnwidth]{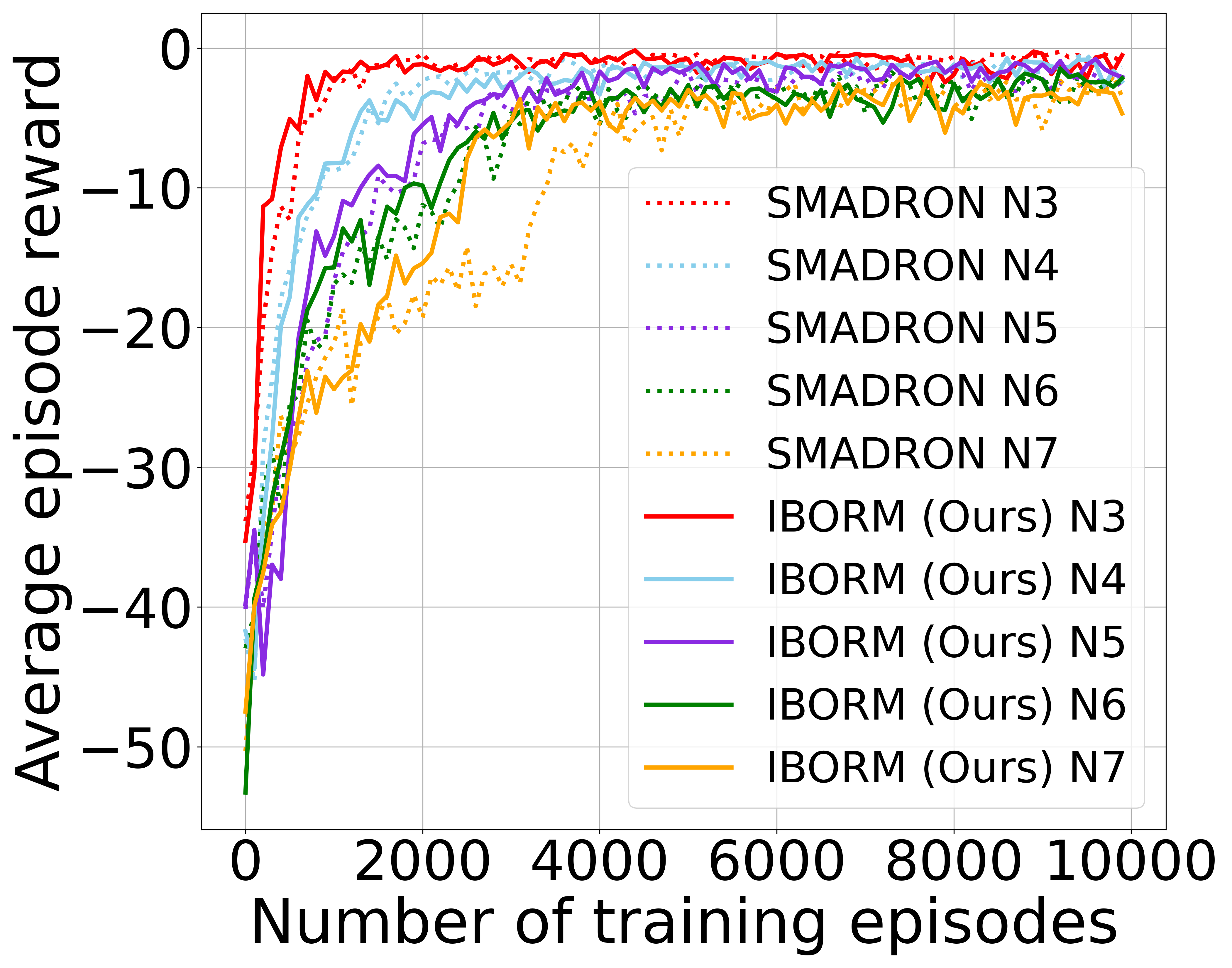}
    \end{minipage}
}
\captionsetup{font=footnotesize}
\caption {
Convergence curves of average episode reward of different methods.
(a) IBORM vs. SMADQN.
(b) IBORM vs. SMADRON.
}
\label{CovergenceCurves}
\end{figure}

We train each method by 10k episodes. At the beginning of each episode, positions of targets and agents are generated randomly.
Each method is evaluated with different numbers of targets and agents ($N=3,4,5,6,7$).
Convergence curves of average episode reward are shown in Fig.\ref{CovergenceCurves}. 
As we can see from the results, when the number of agents increases, IBORM learns faster than the other two methods, which indicates the advantage of IBORM over implicit behavior representation learning (SMADQN) and explicit behavior representation learning without considering information utility (SMADRON).
To test the performance of the learned policies,
we generate $1000$ testing tasks with random positions of targets and agents.
Table \ref{success rate} shows success rate of cooperative navigation with different policies, from which we can see that IBORM outperforms the other two methods consistently. 

\begin{figure}[t]
\centering
\subfigure[]
{
    \begin{minipage}[t]{2.5cm}
	\centering
    \includegraphics[width=1.1\columnwidth]{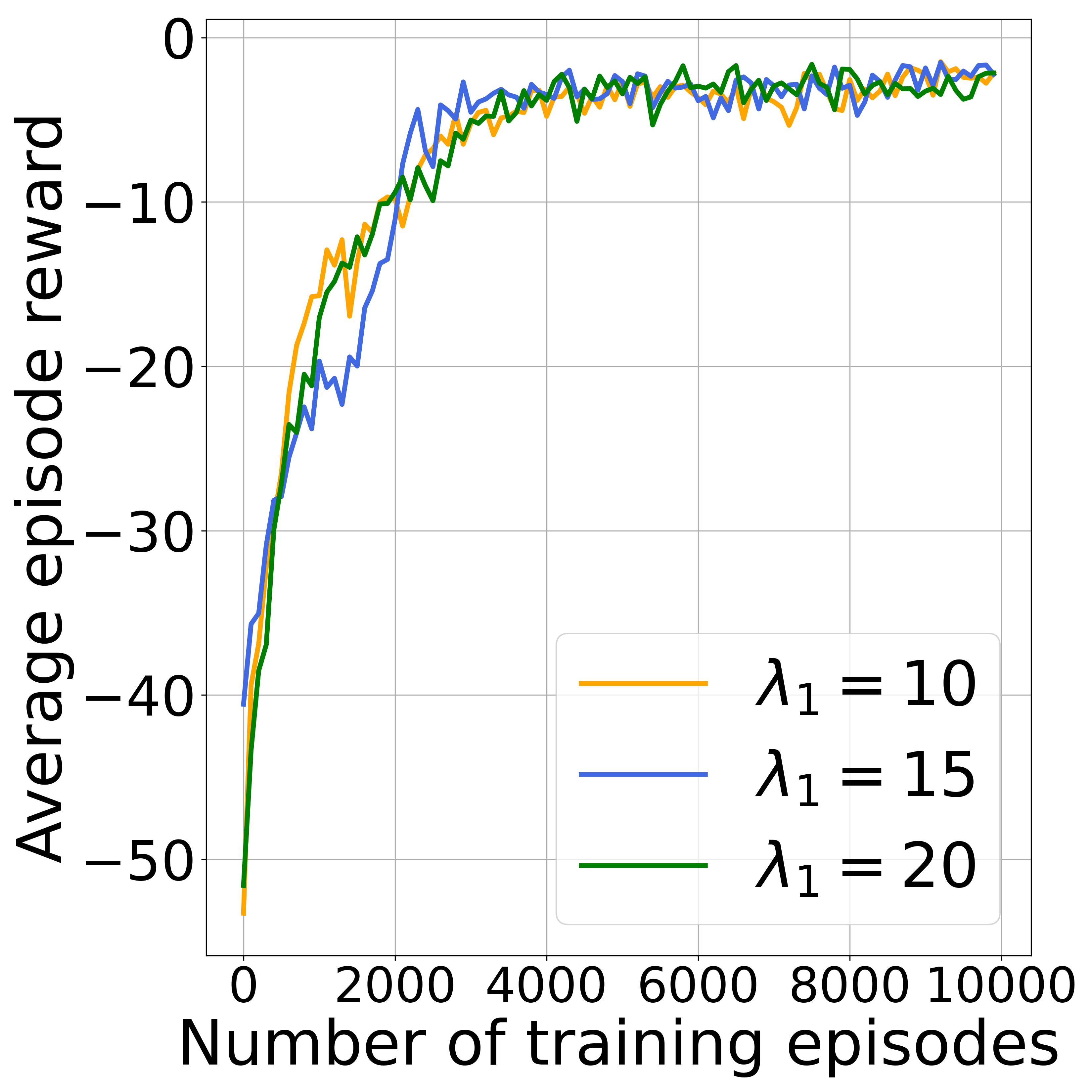}
    \end{minipage}
}
\subfigure[]
{
    \begin{minipage}[t]{2.5cm}
	\centering
    \includegraphics[width=1.1\columnwidth]{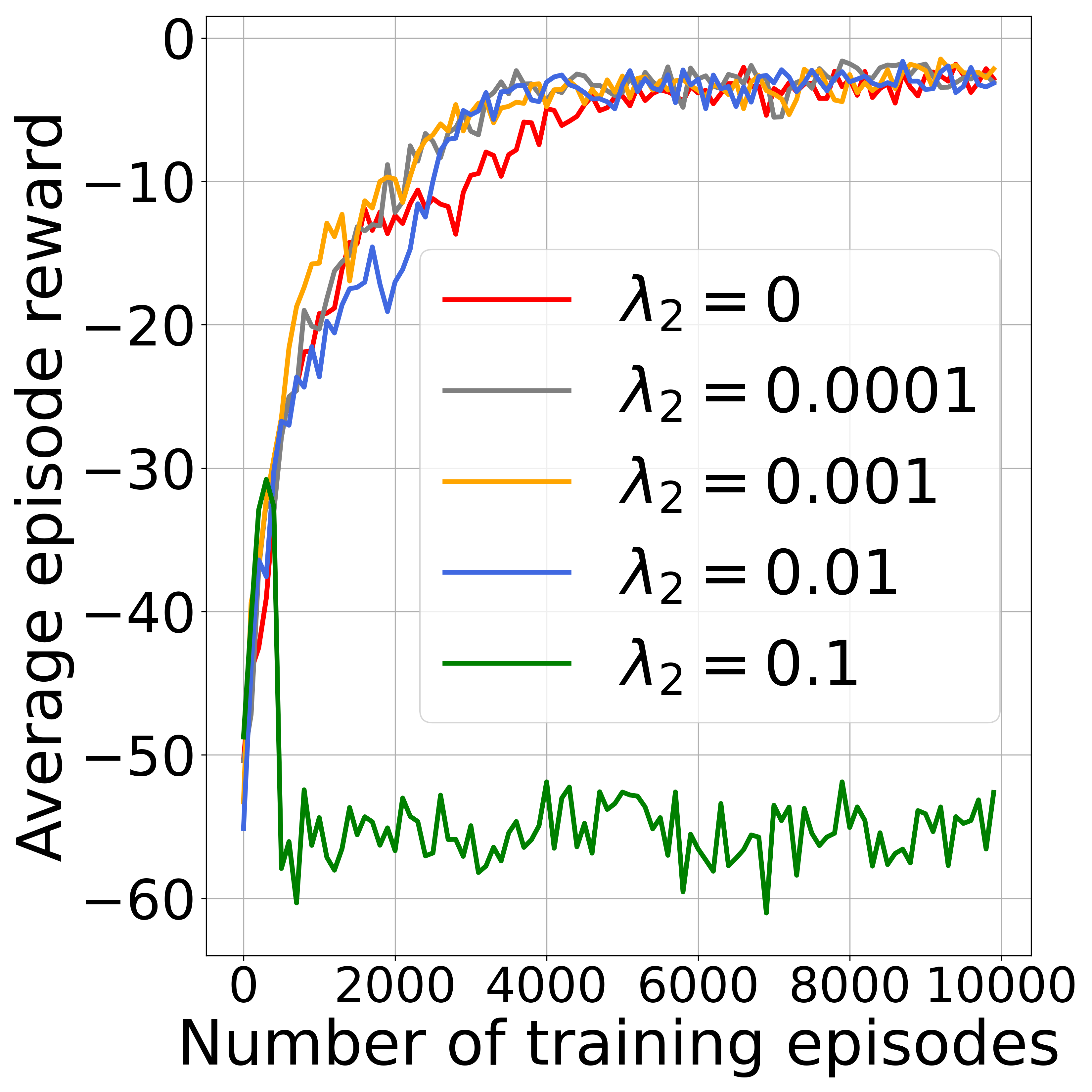}
    \end{minipage}
}
\subfigure[]
{
    \begin{minipage}[t]{2.5cm}
	\centering
    \includegraphics[width=1.1\columnwidth]{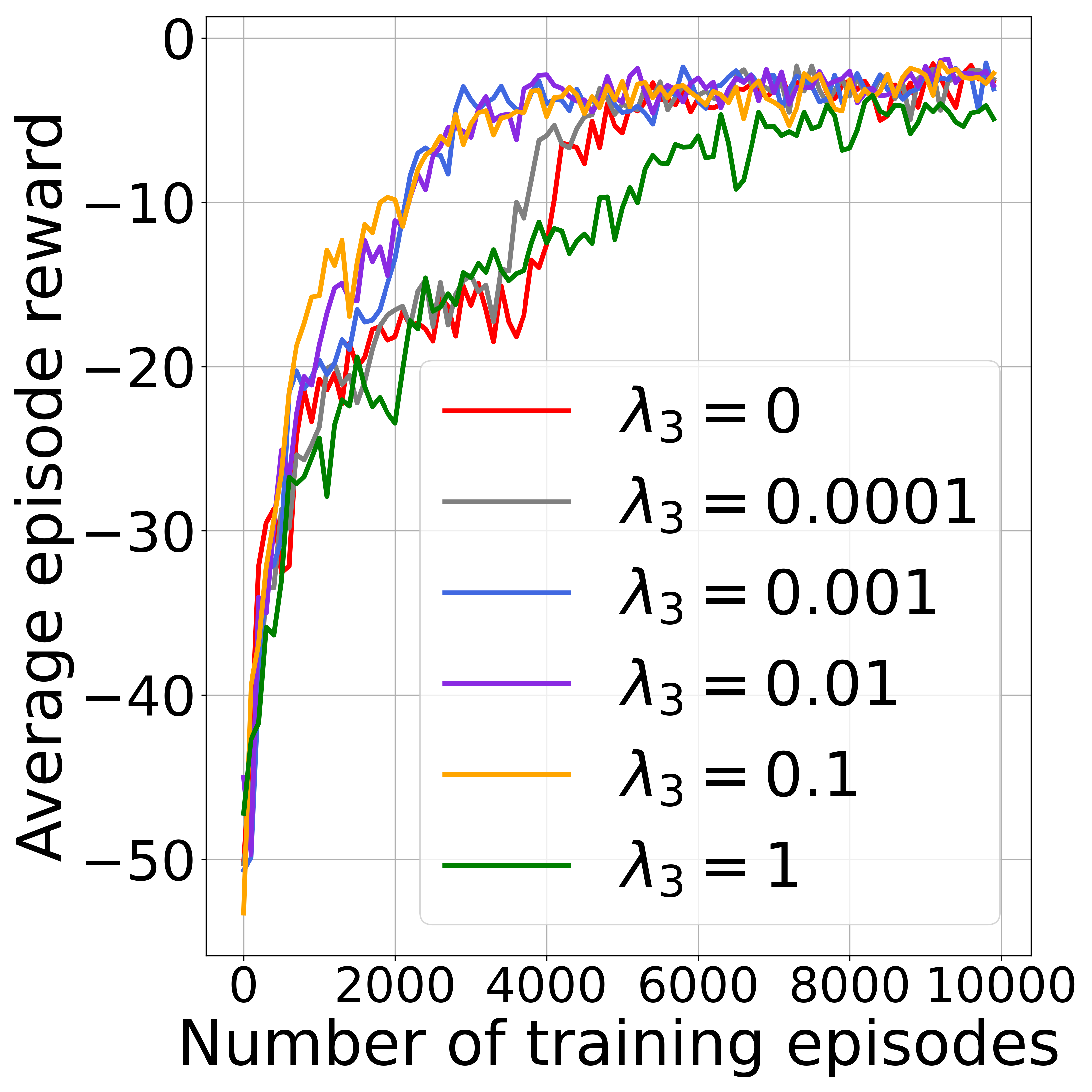}
    \end{minipage}
}
\captionsetup{font=footnotesize}
\caption {
Sensitivity of IBORM to $\lambda$ values on tasks containing six targets and agents. Subfigures (a), (b) and (c) corresponds to $\lambda_1$, $\lambda_2$ and $\lambda_3$, respectively.
}
\label{Lambda}
\end{figure}

We also investigate how learning performance of IBORM changes when $\lambda_1$, $\lambda_2$, and $\lambda_3$ vary in (\ref{loss function_new}).
Fig.\ref{Lambda} shows the results in tasks of $N=6$. Three subfigures corresponds to  $\lambda_1$, $\lambda_2$ and $\lambda_3$, respectively.
As shown in the results, the suitable value range of each $\lambda$ is relatively wide.
Fig.\ref{Lambda}(b) indicates large $\lambda_2$ can lead to failure of IBORM. This is because large $\lambda_2$ causes much information to be discarded.
Fig.\ref{Lambda}(c) shows that large $\lambda_3$ reduces learning speed and gets less rewards, 
because when $\lambda_3$ is enlarged, $\lambda_1$ is weakened relatively and thus the learned representation captures insufficient information regarding the task.
Too small $\lambda_3$ also slows down learning, because the learned representation captures insufficient information about other agents' behaviors and thus provides less help to MADRL. Additionally, when $\lambda_2$ or $\lambda_3$ equals zero, the learning performance degenerates, which indicates the importance of each mutual information constraint in IBORM. 

	\section{Conclusion}
	\label{sec:page}
We propose IBORM to facilitate MADRL by learning a compact and informative representation regarding other agents' behaviors.
We implement IBORM based on our recently proposed MADRL algorithm, SMADQN, by replacing the implicit behavior representation learning of SMADQN with information-bottleneck-based explicit behavior representation learning.
Experimental results demonstrate that IBORM learns faster and the resulting policies can achieve higher success rate consistently, as compared with implicit behavior representation learning (SMADQN) and explicit behavior representation learning (SMADRON) without considering information compression and utility.

\clearpage
\vfill\pagebreak	
\bibliographystyle{IEEEtran}
\bibliography{IBORM}

\end{document}